\documentclass[letterpaper]{article} 
\usepackage{aaai2027}
\nocopyright
\usepackage{amsmath}
\usepackage{multirow}
\usepackage[hyphens]{url}  
\usepackage{graphicx} 
\usepackage{natbib}  
\usepackage{caption} 
\usepackage{algorithm}
\usepackage{algorithmic}
\usepackage{xspace}
\usepackage{graphicx}
\usepackage{subcaption}
\usepackage{newfloat}
\usepackage{listings}
\DeclareCaptionStyle{ruled}{labelfont=normalfont,labelsep=colon,strut=off} 
\floatstyle{ruled}
\newfloat{listing}{tb}{lst}{}
\floatname{listing}{Listing}

\usepackage{booktabs}

\title{A Wrong Turn Does Not Ruin the Journey: Deviation-Guided Skill Self-Evolution for LLM Agents}
\author{
    Yichun Feng\textsuperscript{\rm 1},
    Jiawei Wang\textsuperscript{\rm 2},
    Haozhe Sun\textsuperscript{\rm 3}\corresponding
}
\affiliations{
    \textsuperscript{\rm 1}University of Chinese Academy of Sciences\\
    \textsuperscript{\rm 2}University of Science and Technology of China\\
    \textsuperscript{\rm 3}Meituan\\
    fengyichun22@mails.ucas.ac.cn, wangjiawei@mail.ustc.edu.cn, sunhaozhe02@meituan.com
}

\begin{document}
\newcommand{\ours}{SkillPivot\xspace}
\maketitle

\begin{abstract}
Large language model agents increasingly rely on natural-language skills to solve complex tool-use tasks. However, such tasks often admit multiple valid solution paths, making it inappropriate to improve skills by forcing failed trajectories to match a fixed successful trajectory. Moreover, failed trajectories are rarely entirely wrong: an agent may first collect useful evidence and make meaningful progress, but later deviate into an erroneous suffix. We therefore argue that skill self-evolution should identify where productive problem solving begins to break down, rather than reflect coarsely over the entire failure.
Based on this insight, we propose SkillPivot, a deviation-point-guided framework for skill self-evolution. SkillPivot detects the transition from a useful prefix to an erroneous suffix using execution validity, goal progress, and action diversity. A stronger teacher then continues from the same prefix and produces a successful alternative under the same interaction history. By contrasting the student’s failed suffix with the teacher’s successful suffix, SkillPivot generates localized skill updates while preserving already effective guidance.
Experiments on ToolQA, LogicBench, and WildClawBench show that SkillPivot consistently outperforms competing skill-evolution methods, improves multiple agent models, and produces compact, transferable skill updates.

\end{abstract}


\section{Introduction}

Large language model (LLM) agents have rapidly made personal AI assistants practical in real-world settings, enabling users to complete complex tasks through natural conversation, tool use, and environment interaction~\citep{wu2023autogen,yang2024swe}. By interleaving reasoning with actions, these agents can search the web, query databases, call external APIs, and operate over multi-step workflows. However, their reliability remains limited when task success depends not only on choosing the right tool, but also on following task-specific procedural knowledge: when to query, how to construct tool arguments, how to interpret observations, when to verify intermediate results, and how to recover from unexpected tool feedback~\cite{yao2024tau,xie2024travelplanner}. A common way to provide such procedural knowledge is to equip agents with a skill corpus, where each skill encodes reusable natural-language guidance for task solving and tool use~\citep{wang2023voyager}. Yet even high-quality skills are often incomplete: they may cover the main workflow while omitting boundary cases, recovery strategies, or tool-specific constraints.

\begin{figure}[t]
    \centering
    \includegraphics[width=\linewidth]{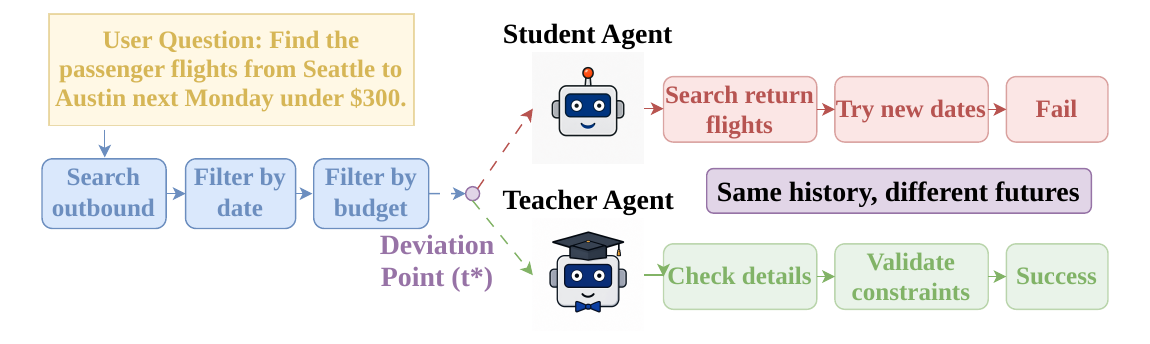}
    \caption{
   A failed agent trajectory can contain a useful prefix before a deviation point, and comparing the divergent suffixes from the same history reveals how the existing skill should be updated.
    }
    \label{fig:motivation}
\end{figure}

This motivates the problem of skill self-evolution: can an agent improve an existing skill corpus from its own interaction failures? Recent work on agent self-improvement and evolving skill libraries suggests that experience can be converted into reusable knowledge. However, many existing approaches rely on whole-trajectory reflection, successful-trajectory imitation, or open-ended rewriting of skill descriptions. These strategies overlook a key property of failed agent trajectories: a failed trajectory is rarely uniformly wrong. In many cases, as illustrated in Figure~\ref{fig:motivation}, the agent first performs useful exploration, gathers relevant evidence, or narrows down the solution space, and only later deviates into repeated invalid tool calls, premature stopping, or incorrect interpretation of observations. Treating the entire trajectory as a single failure signal can therefore mix useful prefix behavior with harmful suffix behavior, leading to noisy, overly broad, or instance-specific skill updates.

We propose \ours{}, a deviation-point-guided framework for skill self-evolution. Instead of rewriting skills from full failed trajectories, \ours{} first detects the step where the student agent begins to deviate from productive problem solving. A stronger teacher agent then continues from the same trajectory prefix, producing a successful suffix whenever possible. By contrasting the student's failed suffix with the teacher's successful suffix, \ours{} identifies localized skill gaps and generates minimal, conditional, and auditable skill deltas. To prevent harmful updates, each candidate delta is further examined by a regression-gated verifier that checks whether the update preserves useful original rules and whether the new guidance generalizes beyond the triggering cases. In this way, \ours{} turns failures into localized maintenance signals rather than uncontrolled skill rewrites.

We evaluate \ours{} on ToolQA using six disjoint task groups, together with additional benchmarks for transfer evaluation. The results show that evolved skills consistently improve performance over the original skill corpus and generalize beyond their source groups, indicating that \ours{} learns reusable procedural corrections rather than source-specific patches. Further analyses show that skill evolution can continue over multiple rounds, that evolved skills can also benefit stronger models, and that detected deviation points are reliable and driven by meaningful signals. Ablation studies further demonstrate the importance of suffix contrast and the role of review in maintaining long-term stability during continuous evolution.


Our contributions are threefold:
\begin{itemize}
\item We find that failures of skill-guided LLM agents are often localized: failed trajectories usually contain useful prefixes before deviating into erroneous suffixes.

\item We propose \ours{}, a deviation-point-guided skill self-evolution framework that uses teacher continuation and suffix contrast to generate minimal, conditional, and regression-checked skill deltas.

\item We empirically show that skill self-evolution produces reusable procedural knowledge rather than instance-specific patches, yielding transferable gains across task groups and evolution rounds.

\end{itemize}

\section{Related Work}
\subsection{Tool-Augmented Language Agents}

Large language models have increasingly been extended into interactive agents that can reason, call tools, and act in external environments. ReAct interleaves reasoning traces with actions, allowing agents to update their plans based on intermediate observations and tool feedback~\citep{yao2022react}. Toolformer shows that language models can learn when and how to invoke external APIs through self-supervised signals~\citep{schick2023toolformer}. Subsequent tool-learning systems, such as ToolLLM and Gorilla, further improve API selection, argument generation, and generalization to unseen or changing tool documentation~\citep{qin2024toolllm,patil2024gorilla}. Meanwhile, benchmarks such as ToolQA, API-Bank, WebArena, Mind2Web, AgentBench, and AgentBoard show that tool-augmented agents still struggle with long-horizon reasoning, grounded tool execution, and reliable task completion in realistic environments~\citep{zhuang2023toolqa,li2023api,zhou2024webarena,deng2023mind2web,liu2024agentbench,ma2024agentboard}. These works establish tool use as a key capability of language agents.

\subsection{Agent Self-Evolution}

Beyond tool use, recent work has explored how agents can improve themselves through experience. One line of work updates model parameters using interaction data, such as supervised fine-tuning on agent trajectories, reinforcement learning from environmental feedback, or preference optimization from human or automatic judgments~\citep{ouyang2022training,bai2022constitutional,yao2022webshop}. These methods can directly change the underlying policy, but they usually require expensive training, carefully curated data, and repeated safety evaluation before deployment.
Another line of work keeps the base model frozen and improves agent behavior through external mechanisms. Reflexion stores verbal feedback from previous trials to guide future decisions~\citep{shinn2023reflexion}. Self-Refine improves outputs through iterative self-feedback and revision~\citep{madaan2023self}. ExpeL extracts reusable natural-language lessons from accumulated experiences~\citep{zhao2024expel}. LATS combines reasoning, acting, reflection, and tree search to improve exploration~\citep{zhou2023language}, while Agent Workflow Memory induces reusable workflows from prior trajectories~\citep{wang2024agent}.

\subsection{Self-Evolving Agent Skills}

A more recent line of work studies agent skills as an evolving layer of procedural knowledge. SkillsBench evaluates whether curated skills can improve LLM-based agents and shows that high-quality skills are useful, while automatically generated skills may be unstable~\citep{li2026skillsbench}. SRA-Bench and related work study how agents retrieve relevant skills from large-scale skill corpora~\citep{su2026skill}. SkillGenBench and SkillLearnBench further evaluate whether agents can generate or continually learn reusable skills from repositories, documents, or previous task experience~\citep{zhou2026skillgenbench,zhong2026skilllearnbench}.
Recent systems further move from static skill libraries to self-evolving skills. SkillWeaver enables web agents to discover and distill website-specific skills~\citep{zheng2025skillweaver}. SkillGen synthesizes auditable skills from successful and failed trajectories~\citep{ma2026skillgen}. SkillForge introduces a creation-evaluation-refinement loop for cloud technical support~\citep{liu2026skillforge}. SkillClaw studies collective skill evolution by aggregating multi-user trajectories and using an agentic evolver to refine or create skills~\citep{ma2026skillclaw}. SkillOpt treats a natural-language skill document as a trainable text-space state and optimizes it through rollouts and validation gates~\citep{yang2026skillopt}.
SkillAudit compares with-skill and without-skill trajectories on the same task to identify effective or harmful content within a skill~\citep{gao2026skillaudit}.
Self-Harness summarizes common issues from multiple failed trajectories and optimizes the overall agent harness through regression validation, rather than an individual skill~\citep{zhang2026self}.
Existing methods typically derive skill-update signals from complete trajectories, paired execution outcomes, or cross-trajectory failure patterns, but may mistakenly treat useful exploration before the failure as behavior that also requires correction, thereby disrupting reusable strategies that are already effective. In contrast, \ours{} identifies the deviation point where a trajectory shifts from productive exploration to erroneous behavior, preserves the useful prefix, and generates a localized skill update only from the erroneous suffix, enabling more precise and less disruptive skill evolution.

\section{Method}

\subsection{Problem Formulation}

We study the problem of skill self-evolution for tool-augmented language agents. Unlike settings that generate skills from scratch, we assume that the agent is equipped with an existing high-quality skill corpus. Let $S^r = \{s_1^r, s_2^r, \ldots, s_M^r\}$ denote the skill corpus at evolution round $r$, where each skill $s_m^r$ is a natural-language procedural artifact that specifies task-solving strategies, tool-use conventions, constraints, and exception-handling rules.

At each evolution round, the agent receives a set of tasks $D^r = \{x_i\}_{i=1}^{N}$. Given a task $x_i$ and the current skill corpus $S^r$, a student agent $A_s$ interacts with the environment and produces a ReAct-style trajectory:
\begin{equation}
\tau_i = (x_i, h_{i,1}, a_{i,1}, o_{i,1}, \ldots, h_{i,T_i}, a_{i,T_i}, o_{i,T_i}, y_i).
\end{equation}
Here, $h_{i,t}$, $a_{i,t}$, and $o_{i,t}$ denote the agent's reasoning state, action or tool call, and environment observation at step $t$, respectively. The final response is denoted by $y_i$. An evaluator $E$ determines whether the trajectory successfully completes the task:
\begin{equation}
E(\tau_i, x_i) \in \{0,1\}.
\end{equation}
The collected trajectories are then partitioned into a success set $T^+$ and a failure set $T^-$.

The goal of skill self-evolution is to learn an update operator $\Phi$ that converts interaction evidence into a set of skill deltas:
\begin{equation}
\Delta S^r = \Phi(S^r, T^+, T^-).
\end{equation}
Applying these deltas to the current skill corpus produces the next-round skill corpus:
\begin{equation}
S^{r+1} = \mathrm{Update}(S^r, \Delta S^r).
\end{equation}
Each skill delta is expected to be minimal, conditional, and auditable. It should repair missing or underspecified procedural knowledge without rewriting the whole skill or overfitting to a single failed instance.

Formally, the desired update should improve task success on future tasks while preserving previously stable behavior:
\begin{equation}
\max_{\Delta S^r}
\; \mathbf{E}_{x \sim D_{\mathrm{future}}}
\left[
E(A_s(x; \mathrm{Update}(S^r, \Delta S^r)), x)
\right].
\end{equation}
This objective is subject to bounded regression on tasks that were already solvable under $S^r$. Therefore, skill self-evolution is not merely a repair problem on past failures, but a constrained maintenance problem: the updated skill corpus should fix historical errors, transfer to unseen tasks, and avoid degrading existing capabilities.

\subsection{Framework Overview}

\begin{figure*}[t]
    \centering
    \includegraphics[width=0.98\textwidth]{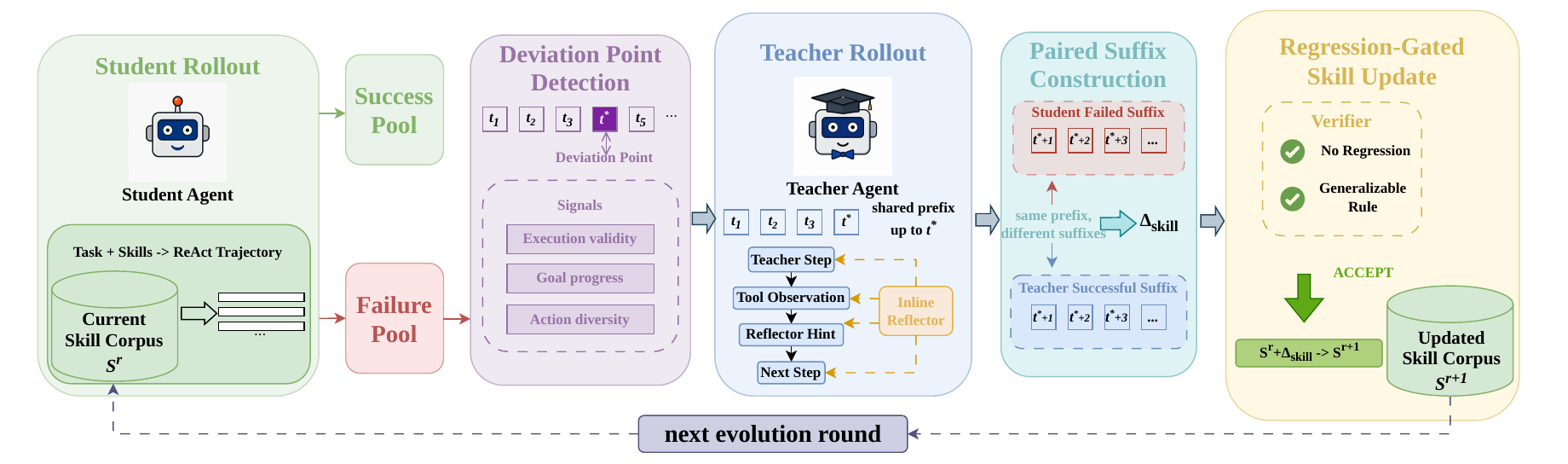}
    \caption{
    Overview of \ours{}.
    }
    \label{fig:framework}
\end{figure*}

We propose a deviation-point-guided framework for skill self-evolution named \ours{}, as illustrated in Figure~\ref{fig:framework}. The key idea is to avoid treating an entire failed trajectory as uniformly wrong. Instead, we first identify the point at which the student agent shifts from useful exploration to unproductive or erroneous behavior, and then use this point as the anchor for skill update. 

The framework consists of five stages. First, the student agent executes tasks with the current skill corpus and produces ReAct-style trajectories. Second, SkillPivot analyzes failed trajectories to locate the deviation point where useful exploration turns into erroneous behavior. Third, a stronger teacher agent continues from the same prefix and attempts to produce a successful suffix. Fourth, SkillPivot contrasts the student failed suffix with the teacher successful suffix to identify the missing procedural guidance and generate candidate skill deltas. Finally, a regression-gated verifier checks whether each delta preserves useful original rules and expresses a generalizable update before merging it into the skill corpus.

This pipeline turns raw interaction failures into localized and auditable skill updates. By preserving the useful prefix of the student trajectory and focusing only on the divergent suffix, the method reduces unnecessary rewriting and makes skill evolution more causally grounded.

\subsection{Student Rollout}

At each evolution round, the student agent is prompted with the current skill corpus and executes a batch of tasks in the environment. Each task produces a multi-step trajectory that records the agent's reasoning, tool calls, observations, and final response. We retain the full interaction trace because many skill-level failures are not visible from the final answer alone. For example, a task may fail because the agent uses a wrong command hierarchy, passes an argument from an incorrect source, ignores an empty tool response, or stops without verifying the environment state.

After execution, an evaluator determines whether each trajectory succeeds. Successful trajectories are stored as a success pool, which is later used to test whether a candidate skill update damages previously working behavior. Failed trajectories are stored as an failure pool, which provides evidence for diagnosing missing or underspecified procedural knowledge.

\subsection{Deviation Point Detection}

Given a failed trajectory, deviation point detection aims to identify the earliest step at which the trajectory shifts from productive exploration to invalid execution or repetitive behavior. The trajectory before this step is retained as the shared prefix for teacher continuation.
For each step $t$, we compute a progress potential score:
\begin{equation}
\hat{V}_t
=
0.4\,\mathrm{Exec}_t
+
0.3\,\mathrm{Prog}_t
+
0.3\,\mathrm{Div}_t.
\end{equation}
Here, $\mathrm{Exec}_t$ measures whether the current tool call is executed successfully. It is computed using deterministic rules based on action validity, tool execution status, and the usability of the returned observation. Successful executions with usable results receive higher scores, whereas invalid calls, execution errors, and empty observations receive lower scores.
$\mathrm{Prog}_t$ measures the relevance of the current observation to the task objective. We use the ground-truth as a semantic representation of the task objective. The all-MiniLM-L6-v2\cite{reimers-2019-sentence-bert} is used to encode the current observation and the ground-truth, after which their cosine similarity is computed. A higher similarity indicates that the observation contains information more closely related to the target information required by the task. To prevent repeated evidence from being counted multiple times as progress, we retain only the positive increase in target relevance over the best previous step:
\begin{equation}
\mathrm{Prog}_t
=
\max
\left(
0,\,
\operatorname{sim}(o_t,y)
-
\max_{j<t}\operatorname{sim}(o_j,y)
\right),
\end{equation}
where $y$ denotes the ground-truth. For the first step, the historical maximum is defined as zero. Therefore, an observation receives a positive progress score only when its relevance to the target answer exceeds that of all previous observations. Repeated evidence does not produce additional progress.
$\mathrm{Div}_t$ measures whether the current action repeats a previously attempted strategy. We use the all-MiniLM-L6-v2 to encode the actions and define action diversity as
\begin{equation}
\mathrm{Div}_t
=
1
-
\max_{j<t}
\operatorname{sim}(a_t,a_j).
\end{equation}
A low value of $\mathrm{Div}_t$ indicates that the current action is highly similar to a previous action and may represent repetitive behavior, whereas a high value indicates a meaningfully different strategy.

After obtaining the score for each step, the detector first treats the step with the lowest $\hat{V}_t$ as the candidate deviation point. To avoid mistaking a single accidental low-scoring step for a true deviation, we further examine whether the progress potential after this step remains consistently lower than the overall level before it. The candidate is confirmed as the deviation point only when it forms a clear trough in the trajectory and the subsequent steps continue to maintain low scores. This design requires the trajectory to exhibit sustained invalid execution and clear repetitive behavior, thereby preventing premature truncation of prefixes that may still contain useful exploration.

\subsection{Teacher Rollout}

Once a reliable deviation point is detected, SkillPivot asks a stronger teacher agent to continue from the same prefix rather than restarting the task from scratch. The teacher receives the task, the current skill corpus, the student's pre-deviation trajectory prefix, and the student's full failed trajectory as a reference of what to avoid. The teacher then continues the ReAct loop from step $t^*$ with the same step numbering and interacts with the environment through valid tool calls.

To improve the quality of teacher continuation, the implementation uses an inline reflector. After each teacher tool-call step, the reflector analyzes the latest action and observation and produces a short hint for the next step. This hint is appended to the teacher scratchpad as a special reflector message, but it is not inserted into the saved raw teacher trajectory. Thus, the reflector can guide the teacher online without polluting the suffix that will later be used for comparison.
A teacher continuation is accepted only if it passes the task evaluator. If the teacher succeeds, SkillPivot constructs a paired case consisting of the shared prefix, the student's failed suffix, and the teacher's successful suffix. If the teacher fails, the case is discarded or retried from the same deviation point up to a preset attempt limit.

\subsection{Paired Suffix Contrast and Skill Generation}

After a teacher continuation is verified as successful, SkillPivot constructs a paired suffix case. Each case contains the task, the current skill, the detected deviation point, the student's failed suffix after $t^*$, the teacher's successful suffix from the same prefix, and lightweight difference statistics.
First, a single-case trajectory summarizer analyzes each teacher-correct paired case. It identifies the student's mistake, the teacher's correction, the missing or incomplete guidance in the current skill, and a general evolution direction. This produces a structured case analysis rather than an immediate skill rewrite.
Second, a skill increment generator aggregates the structured analyses and proposes a candidate delta. The generator can choose one of three actions: \textit{improve\_skill}, which updates the skill content; \textit{optimize\_description}, which refines the skill description or triggering condition; or \textit{skip}, which avoids editing when the evidence is insufficient. 

\subsection{Regression-Gated Skill Update}

The generated skill delta is not directly deployed as the final skill. Instead, SkillPivot first uses an LLM editor to apply the candidate delta to the original skill and produce a tentative updated skill.
Since the original skill corpus already contains useful procedural knowledge, the tentative updated skill $s'$ may still introduce harmful changes, such as deleting valid rules, weakening existing guidance, or adding instance-specific patches. Therefore, before deploying $s'$, SkillPivot applies a regression-gated verifier to decide whether the updated skill can be safely merged into the skill corpus.

The verifier is an LLM-based reviewer. Given the original skill $s$ and the tentative updated skill $s'$, the verifier compares their contents using two criteria. The first criterion, \textit{no regression}, checks whether $s'$ preserves useful rules from $s$ and does not delete, weaken, override, or contradict them without justification. The second criterion, \textit{generalizable rule}, checks whether the newly added guidance expresses a reusable procedural principle rather than a one-off patch tailored to the triggering failure cases.
The verifier assigns both criteria a score in $[0,1]$ and accepts the update only when both scores exceed a threshold:
\begin{equation}
r_{\mathrm{reg}} \geq \theta
\quad \text{and} \quad
r_{\mathrm{gen}} \geq \theta .
\end{equation}
In our implementation, $\theta=0.5$. If accepted, $s'$ is written into the deployed updated corpus and used in the next evolution round. If rejected, the original skill $s$ remains unchanged, and the delta is retained only as a candidate record. This gate makes skill self-evolution conservative: each deployed update must repair observed failures without sacrificing previously useful skill behavior.

\section{Experiments}

\subsection{Experimental Setup}
Unless otherwise specified, we use Qwen3-32B~\cite{qwen3technicalreport} as the student agent to execute tasks and generate final answers, while all LLM-based modules in SkillPivot, including teacher continuation, trajectory summarization, skill generation, and regression-gated verification, are instantiated with Qwen3.5-397B-A17B~\cite{qwen3.5}. For both student and teacher agent inference, we set the sampling temperature to 0.7 and the maximum generation length to 8,096 tokens per LLM call. The student ReAct engine is allowed to execute at most 20 steps per task, while the teacher ReAct engine is allowed to execute at most 10 continuation steps from the detected deviation point and can be retried up to 3 times if continuation fails. For self-evolution modules, the default maximum generation length is set to 12,000 tokens. For the cross-model transfer experiment, the evaluated models are explicitly specified in the corresponding section. All reported task-performance metrics follow the official evaluation protocol of each benchmark.



\subsection{RQ1: Do Evolved Skills Improve Performance and Generalize Beyond Their Source Groups?}

We conduct this experiment on ToolQA~\cite{zhuang2023toolqa}, using the skill corpus configured by SRA-Bench~\cite{su2026skill} and splitting the test set into six disjoint groups, with 238 examples per group. The Baseline row evaluates the original skill corpus without self-evolution on all groups. Each \ours{}-g$i$ row denotes the cumulatively evolved skill corpus after incorporating evolution trajectories up to group g$i$, which is then evaluated on all six target groups. All configurations are independently run three times.

As shown in Table~\ref{tab:effectiveness_transfer}, evolved skills consistently outperform the Non-evo across all test groups. The final evolved corpus, \ours{}-g5, improves the mean success rate from 53.18\% to 59.36\%. More importantly, the improvements are not limited to the groups already used during evolution. The table shows consistent gains across all target groups, indicating that cumulative skill evolution improves the skill corpus as a whole rather than only repairing isolated source-group failures.
These results suggest that \ours{} does not simply memorize previously observed failures or produce narrow group-specific patches. Instead, the evolved skills capture recurring failure patterns shared across task groups and convert them into reusable local rules. Therefore, this experiment supports two conclusions: skill evolution improves overall task performance, and the resulting skills continue to generalize as the skill corpus is cumulatively updated.

\begin{table}[t]
\centering
\small
\resizebox{\linewidth}{!}{
\begin{tabular}{lccccccc}
\toprule
\textbf{Corpus}
& \textbf{g0(\%)}
& \textbf{g1(\%)}
& \textbf{g2(\%)}
& \textbf{g3(\%)}
& \textbf{g4(\%)}
& \textbf{g5(\%)}
& \textbf{Mean(\%)} \\
\midrule
Non-evo
& 52.59$\pm$0.44
& 53.14$\pm$1.05
& 50.98$\pm$0.48
& 54.20$\pm$0.42
& 53.51$\pm$1.03
& 54.63$\pm$0.21
& 53.18$\pm$0.32 \\
\midrule
\ours{}-g0
& 56.08$\pm$0.63
& 54.26$\pm$0.43
& 52.10$\pm$0.00
& 56.86$\pm$0.64
& 54.21$\pm$0.55
& 54.78$\pm$0.86
& 54.72$\pm$0.21 \\
\ours{}-g1
& 55.88$\pm$0.42
& 54.62$\pm$0.00
& 53.92$\pm$0.24
& 57.00$\pm$0.24
& 55.75$\pm$0.84
& 54.77$\pm$0.31
& 55.32$\pm$0.14 \\
\ours{}-g2
& 55.80$\pm$1.39
& 56.36$\pm$1.15
& 56.02$\pm$0.64
& 58.54$\pm$0.24
& 58.43$\pm$0.63
& 58.85$\pm$1.48
& 57.33$\pm$0.22 \\
\ours{}-g3
& 56.50$\pm$0.38
& 58.18$\pm$0.91
& 57.98$\pm$1.11
& 59.24$\pm$0.42
& 54.49$\pm$0.52
& 57.86$\pm$0.51
& 57.38$\pm$0.21 \\
\ours{}-g4
& 57.62$\pm$0.83
& \textbf{58.74$\pm$1.27}
& \textbf{58.96$\pm$0.48}
& \textbf{61.76$\pm$1.12}
& 56.46$\pm$0.72
& 54.78$\pm$1.08
& 58.05$\pm$0.42 \\
\ours{}-g5
& \textbf{60.42$\pm$0.44}
& 58.18$\pm$1.71
& 58.26$\pm$1.47
& 60.36$\pm$0.24
& \textbf{59.41$\pm$0.72}
& \textbf{59.55$\pm$0.51}
& \textbf{59.36$\pm$0.20} \\
\bottomrule
\end{tabular}
}
\caption{
Effectiveness and cross-group generalization of evolved skills.
}
\label{tab:effectiveness_transfer}
\end{table}


\subsection{RQ2: Does \ours{} Outperform Other Skill Evolution Methods?}

\begin{figure}[t]
    \centering
    \begin{minipage}[t]{0.50\linewidth}
        \centering
        \includegraphics[width=\linewidth]{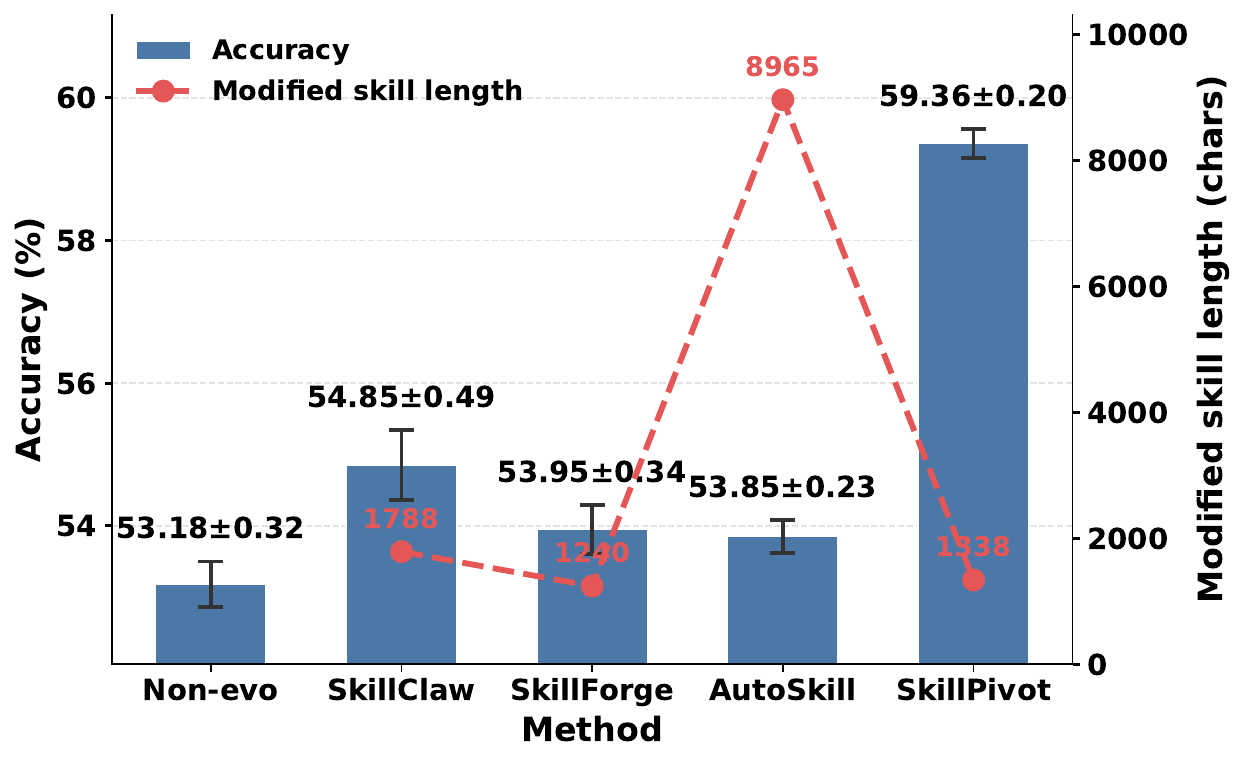}
        \centerline{(a)}
    \end{minipage}
    \hfill
    \begin{minipage}[t]{0.46\linewidth}
        \centering
        \includegraphics[width=\linewidth]{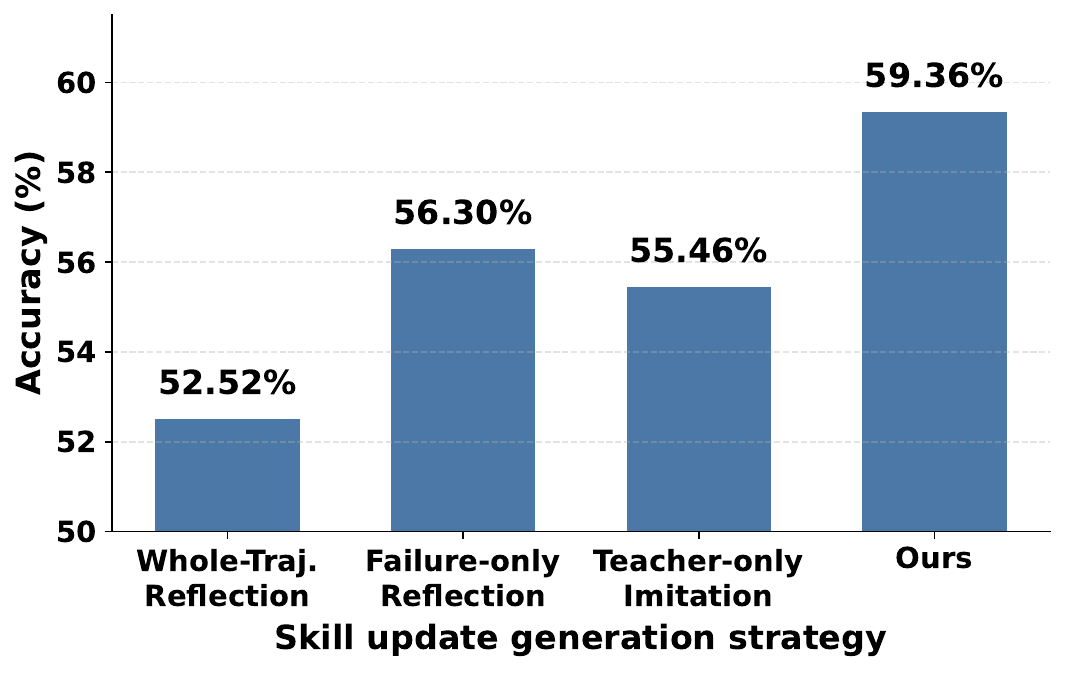}
        \centerline{(b)}
    \end{minipage}
    \caption{
(a) Comparison with other skill evolution methods. (b) Comparison of skill update generation strategies.
    }
    \label{fig:method_comparison}
\end{figure}

We compare \ours{} with three other skill evolution methods: SkillClaw, SkillForge, and AutoSkill\cite{yang2026autoskill}. All methods are applied to the same ToolQA skill corpus and evaluated under the same downstream setting after incorporating their generated skill updates,  as shown in Figure~\ref{fig:method_comparison}(a). All configurations are independently run three times. 
The results show that skill evolution is not simply a matter of making more edits. AutoSkill makes the largest modifications. It tends to rewrite the original skill into a more general prompt-style template, such as adding role, objective, and general instruction components. Although this type of rewriting makes the skill longer and more complete, a substantial portion of the added content is not directly related to specific retrieval failures or tool-use failures. As a result, it may dilute the truly critical operational rules and ultimately brings only limited improvement. SkillClaw and SkillForge make lighter edits, mainly adding local constraints or heuristic rules to the original skill, such as date normalization, distinction between score semantics, query field restrictions, or hard retrieval constraints. These edits can fix some local errors, but they usually lack a systematic recovery procedure for retrieval failures, tool-state issues, field ambiguity, and answer-format errors. Therefore, their improvements remain limited.

In contrast, \ours{} achieves the highest accuracy with a relatively compact modified skill length. Its updates convert observed failures into executable recovery protocols, such as retrying with alternative date formats, relaxing overly specific constraints, reformulating queries, and verifying candidate evidence when retrieval fails. This suggests that effective skill evolution depends less on update length than on whether failures are transformed into reusable procedural guidance.

\subsection{RQ3: Do Evolved Skills Also Improve Other Models?}

To examine whether evolved skills are model-specific patches or transferable execution knowledge, we evaluate the original and evolved skill corpora with five different models: GPT-4o mini\cite{openai2024gpt4omini}, Gemini 3.1 Flash-Lite\cite{google2026gemini31flashlite}, DeepSeek-V4-Pro\cite{xu2026deepseek}, Qwen3.5-397B-A17B, and Qwen3-32B. All models are evaluated on the same six task groups. The original skill corpus is used as the baseline, while the one-round evolved corpus is used as the evolved setting.

\begin{figure}[t]
    \centering
    \begin{minipage}[t]{0.55\linewidth}
        \centering
        \includegraphics[width=\linewidth]{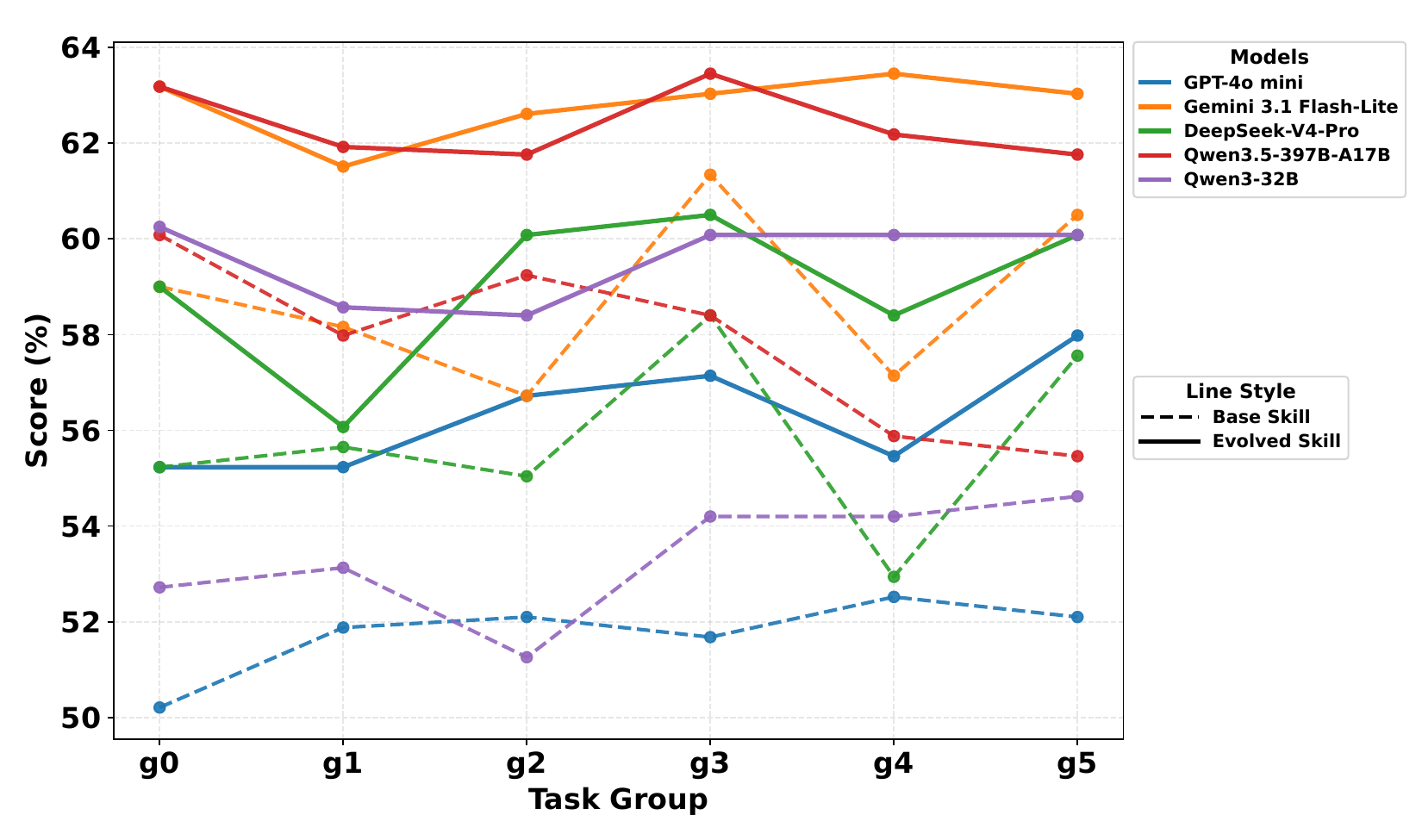}
        \centerline{(a)}
    \end{minipage}
    \hfill
    \begin{minipage}[t]{0.41\linewidth}
        \centering
        \includegraphics[width=\linewidth]{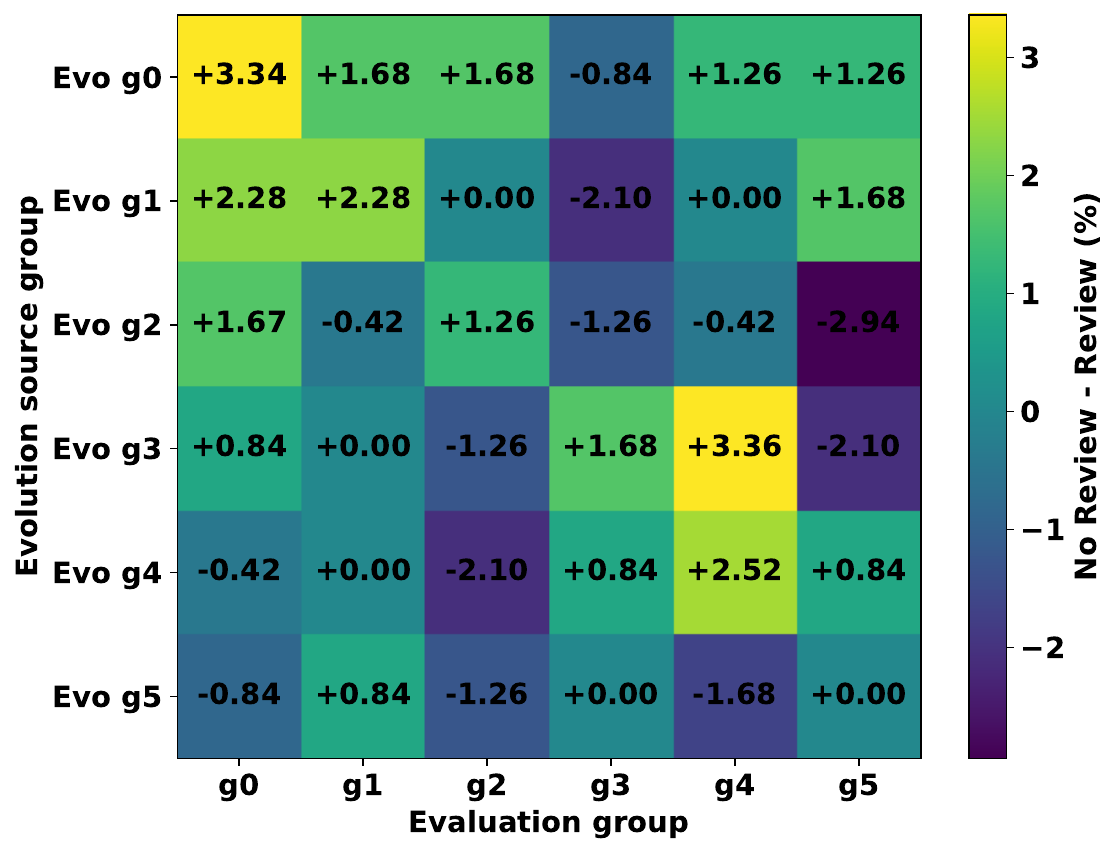}
        \centerline{(b)}
    \end{minipage}
    \caption{
(a) Cross-model transfer of evolved skills on ToolQA. (b) Review ablation on ToolQA. Values denote no-review minus review accuracy.
    }
    \label{fig:cross_model_and_review}
\end{figure}

As shown in Figure~\ref{fig:cross_model_and_review}(a), the evolved skill corpus consistently improves performance across all evaluated models. The improvement is not limited to the model that generated the skill updates, nor is it restricted to a small subset of task groups. Instead, the evolved skills lead to positive gains for every model, suggesting that the updates capture reusable task-execution knowledge rather than model-specific corrections.
More importantly, some weaker models equipped with the evolved skill corpus become competitive with, and in some cases surpass, stronger models using the original skill corpus. This suggests that skill self-evolution does not merely repair isolated failures for a particular model. Instead, it distills clearer and more executable task strategies into the external skill corpus.

\subsection{RQ4: Are Detected Deviation Points Reliable and Driven by Meaningful Signals?}

We evaluate whether the deviation points detected by \ours{} identify reliable and useful intervention positions on ToolQA. From Scratch does not use failed trajectory information and directly lets the teacher re-execute the task from the beginning. Random Intervention randomly selects an intermediate step from the failed trajectory as the continuation point. Direct LLM Locator lets an LLM read the complete failed trajectory once and directly output the deviation point it considers most reasonable. Step-wise LLM Verifier performs step-by-step judgments over the trajectory and selects the position most likely to indicate the deviation point. The self-teacher ablation uses the deviation point detected by \ours{}, but replaces the teacher with the same Qwen3-32B model as the student, in order to disentangle the effect of deviation localization from teacher model capability. For methods that select an intervention point, we use three independent LLM judges, DeepSeek-V4-Pro, Qwen3.5-397B-A17B, and Claude Opus 4.6~\cite{anthropic2026claudeopus46}, to assess whether the selected point marks a reasonable transition from useful exploration to erroneous deviation. Point Agree. is computed by majority vote among the three judges. We further let the teacher continue from the corresponding position and measure continuation success, average continuation steps, and invalid tool-call rate.

As shown in Table~\ref{tab:deviation_intervention}, Random Intervention slightly improves over From Scratch, suggesting that failed trajectories do contain reusable intermediate information. However, its Point Agree. is very low, and its Avg. Steps and Invalid Calls are significantly higher, indicating that random points often fail to correspond to the true turning point of failure. LLM-based methods achieve high Point Agree., but their continuation success remains lower than \ours{}, suggesting that a deviation point that appears reasonable is not necessarily the most useful for downstream recovery. In the self-teacher ablation, \ours{} still maintains high Point Agree. and low Invalid Calls, indicating that deviation localization does not fully depend on a stronger teacher. However, the weaker teacher substantially reduces continuation success, showing that teacher capability affects recovery success. Overall, the advantage of \ours{} lies in more accurately locating the boundary between the useful prefix and the erroneous suffix, thereby providing a more effective starting point for teacher continuation and subsequent skill update generation.

\begin{table}[t]
\centering
\small
\resizebox{\linewidth}{!}{
\begin{tabular}{llcccc}
\toprule
\textbf{Method}
& \textbf{Teacher}
& \textbf{Point Agree.} 
& \textbf{Cont. Success}
& \textbf{Avg. Steps} 
& \textbf{Invalid Calls} \\
\midrule
From Scratch
& Qwen3.5-397B
& --
& 35.65\%
& 6.57
& 4.18\% \\

Random Intervention
& Qwen3.5-397B
& 13.32\%
& 38.06\%
& 10.98
& 29.68\% \\

\ours{}
& Qwen3-32B
& 70.81\%
& 37.40\%
& 11.71
& 3.15\% \\

Direct LLM Locator
& Qwen3.5-397B
& 70.57\%
& 45.28\%
& 8.10
& 3.66\% \\

Step-wise LLM Verifier
& Qwen3.5-397B
& \textbf{72.18}\%
& 49.55\%
& 8.08
& 3.45\% \\

\ours{}
& Qwen3.5-397B
& 70.81\%
& \textbf{56.18\%}
& \textbf{5.95}
& \textbf{2.93\%} \\

\bottomrule
\end{tabular}
}
\caption{
Reliability and usefulness of different intervention strategies.
}
\label{tab:deviation_intervention}
\end{table}

\subsection{RQ5: Does Paired Suffix Contrast Produce Better Skill?}

Figure~\ref{fig:method_comparison}(b) isolates the effect of paired suffix contrast in skill generation by comparing different forms of trajectory evidence used to produce skill updates. Unlike RQ2, which compares complete skill evolution methods, this experiment focuses on whether contrasting the failed student suffix with the successful teacher suffix from the same deviation point provides better evidence for generating precise skill. Whole-Trajectory Reflection performs worse than the original skill corpus, suggesting that reflecting on the entire failed trajectory can introduce noisy or overly broad updates. Since the full trajectory contains both useful early steps and later failure behavior, a holistic reflection may fail to isolate the actual point where the skill should be revised.
Failure-only Reflection and Teacher-only Imitation both improve downstream accuracy, but their gains are limited. Failure-only Reflection observes only what the student did wrong, without access to a successful alternative under the same task context. Teacher-only Imitation observes a successful trajectory, but does not explicitly identify which part of the student's behavior caused the failure. As a result, both strategies provide incomplete evidence for generating precise skill deltas.
In contrast, \ours{} achieves the best downstream accuracy. Notably, \ours{} also produces the shortest skill updates on average. This indicates that paired suffix contrast does not improve performance by generating longer or more verbose revisions. Instead, it compares the failed student suffix with the successful teacher suffix from the same deviation point, allowing the update generator to localize the key behavioral difference and produce compact, targeted, and effective skill deltas.


\subsection{RQ6: Does Review Improve Long-Term Stability of Skill Evolution?}

We evaluate the review mechanism on ToolQA using the same six-group split as in previous experiments, where skill deltas are evolved from each source group and evaluated on all target groups. Figure~\ref{fig:cross_model_and_review}(b) shows the ablation results, where each value denotes the accuracy difference between the no-review and review settings. Overall, the no-review setting achieves slightly higher short-term accuracy in some task groups. This suggests that, at the early stage of self-evolution, overly strict review or filtering may discard some potentially useful skill deltas, thereby limiting short-term performance gains. However, this does not imply that the review mechanism is unnecessary. As self-evolution continues, skill deltas gradually accumulate, and the risks of noisy updates, overfitted rules, and incorrect failure attribution become increasingly amplified. In such a long-term evolving system, the role of review is not to maximize the average score in every individual setting, but to suppress unreliable deltas and reduce worst-case degradation and negative transfer. Therefore, the review mechanism should be viewed as a long-term stability constraint rather than a module that always improves short-term accuracy. In other words, the purpose of review is not to improve the average accuracy in every run, but to maintain update quality during continuous skill self-evolution and prevent the system from gradually degrading due to the accumulation of erroneous or overfitted skill.




\subsection{RQ7: Does \ours{} Transfer to Other Benchmarks?}

We further evaluate \ours{} on LogicBench~\cite{parmar2024logicbench} and WildClawBench~\cite{Ding_WildClawBench} to assess its cross-task transferability and continual evolution capability. All configurations are independently run three times.

On LogicBench, we use the skill corpus configured by SRA-Bench~\cite{su2026skill} and evenly divide the examples into six groups. Each Evo \(g_i\) setting evolves the skills using group \(g_i\) and is then evaluated over all six groups. As shown in Table~\ref{tab:cross_group_summary}, \ours{} consistently achieves the best performance across different source-group settings. Since the improvements extend beyond the group used for evolution, the generated updates capture reusable procedural knowledge rather than group-specific corrections. In contrast, the competing methods exhibit smaller or less stable gains, suggesting that unconstrained skill modification may introduce noisy or overly specific rules.

On WildClawBench, we evaluate all tasks that provide and invoke skills. Each Evo \(g_i\) represents one round of continuous evolution over the complete task set, followed by evaluation after that round. As shown in Table~\ref{tab:cross_group_WildClawBench}, \ours{} improves consistently as evolution proceeds and maintains a clear advantage over the competing methods. This indicates that its localized updates can accumulate effectively across rounds without being dominated by redundant or harmful revisions.

\begin{table}[t]
\centering
\scriptsize
\setlength{\tabcolsep}{2.2pt}
\renewcommand{\arraystretch}{1.08}
\resizebox{\linewidth}{!}{
\begin{tabular}{lc|cccccc}
\toprule
\textbf{Method}
& \textbf{Non-evo(\%)}
& \textbf{Evo g0(\%)}
& \textbf{Evo g1(\%)}
& \textbf{Evo g2(\%)}
& \textbf{Evo g3(\%)}
& \textbf{Evo g4(\%)}
& \textbf{Evo g5(\%)}
\\
\midrule

SkillCLAW
& \multirow{4}{*}{\(86.54{\pm}1.05\)}
& \(87.46{\pm}1.27\)
& \(87.58{\pm}0.47\)
& \(87.45{\pm}0.48\)
& \(87.14{\pm}0.40\)
& \(88.02{\pm}0.28\)
& \(88.15{\pm}0.75\)
\\

SkillForge
&
& \(87.19{\pm}0.31\)
& \(88.07{\pm}0.77\)
& \(87.80{\pm}0.06\)
& \(88.32{\pm}0.50\)
& \(89.50{\pm}0.92\)
& \(89.22{\pm}0.61\)
\\

AutoSkill
&
& \(86.30{\pm}0.47\)
& \(87.27{\pm}0.82\)
& \(85.91{\pm}1.46\)
& \(85.52{\pm}1.59\)
& \(82.00{\pm}2.43\)
& \(85.55{\pm}0.79\)
\\

\ours{}
&
& \(\textbf{88.01}{\pm}1.32\)
& \(\textbf{88.80}{\pm}1.32\)
& \(\textbf{88.01}{\pm}1.35\)
& \(\textbf{89.20}{\pm}1.54\)
& \(\textbf{90.83}{\pm}0.80\)
& \(\textbf{91.22}{\pm}1.10\)
\\

\bottomrule
\end{tabular}
}
\caption{
Group-level transfer results on LogicBench.
}
\label{tab:cross_group_summary}
\end{table}

\begin{table}[t]
\centering
\scriptsize
\setlength{\tabcolsep}{2.2pt}
\renewcommand{\arraystretch}{1.08}
\resizebox{\linewidth}{!}{
\begin{tabular}{lc|cccccc}
\toprule
\textbf{Method}
& \textbf{Non-evo(\%)}
& \textbf{Evo g0(\%)}
& \textbf{Evo g1(\%)}
& \textbf{Evo g2(\%)}
& \textbf{Evo g3(\%)}
& \textbf{Evo g4(\%)}
& \textbf{Evo g5(\%)}
\\
\midrule

SkillCLAW
& \multirow{4}{*}{\(34.35{\pm}0.68\)}
& \(34.83{\pm}0.61\)
& \(39.32{\pm}0.88\)
& \(43.68{\pm}1.34\)
& \(55.81{\pm}2.73\)
& \(60.11{\pm}1.48\)
& \(65.27{\pm}0.96\)
\\

SkillForge
&
& \(32.17{\pm}1.90\)
& \(34.71{\pm}1.16\)
& \(43.74{\pm}1.47\)
& \(51.12{\pm}0.57\)
& \(55.17{\pm}0.99\)
& \(60.73{\pm}0.92\)
\\

AutoSkill
&
& \(34.37{\pm}1.65\)
& \(38.91{\pm}0.93\)
& \(41.87{\pm}0.46\)
& \(55.03{\pm}2.36\)
& \(58.83{\pm}1.40\)
& \(63.45{\pm}0.87\)
\\

\ours{}
&
& \(\textbf{36.36}{\pm}0.83\)
& \(\textbf{52.81}{\pm}2.17\)
& \(\textbf{57.93}{\pm}1.37\)
& \(\textbf{64.58}{\pm}1.41\)
& \(\textbf{70.40}{\pm}1.69\)
& \(\textbf{73.00}{\pm}2.55\)
\\

\bottomrule
\end{tabular}
}
\caption{
Group-level transfer results on WildClawBench.
}
\label{tab:cross_group_WildClawBench}
\end{table}








\section{Conclusion}

We study skill self-evolution for LLM agents, where the goal is to improve an existing skill corpus from interaction failures without overwriting previously useful procedural knowledge. Our key observation is that failed trajectories are often not uniformly wrong: they usually contain useful exploratory prefixes before deviating into localized erroneous suffixes. Based on this observation, we propose \ours{}, a deviation-point-guided framework that detects where failures begin, uses teacher continuation to construct successful suffixes, and generates minimal, conditional, and regression-checked skill deltas through suffix contrast. Experiments show that evolved skills produce reusable procedural knowledge rather than instance-specific patches, leading to transferable improvements across task groups, evolution rounds, agent backbones, and task environments. These results suggest that maintaining and evolving external skill corpora is a promising path toward more reliable and continuously improving agents.

\bibliography{aaai2027}

@article{yao2022react,
  title={React: Synergizing reasoning and acting in language models},
  author={Yao, Shunyu and Zhao, Jeffrey and Yu, Dian and Du, Nan and Shafran, Izhak and Narasimhan, Karthik and Cao, Yuan},
  journal={arXiv preprint arXiv:2210.03629},
  year={2022}
}

@article{schick2023toolformer,
  title={Toolformer: Language models can teach themselves to use tools},
  author={Schick, Timo and Dwivedi-Yu, Jane and Dess{\`\i}, Roberto and Raileanu, Roberta and Lomeli, Maria and Hambro, Eric and Zettlemoyer, Luke and Cancedda, Nicola and Scialom, Thomas},
  journal={Advances in neural information processing systems},
  volume={36},
  pages={68539--68551},
  year={2023}
}

@inproceedings{qin2024toolllm,
  title={Toolllm: Facilitating large language models to master 16000+ real-world apis},
  author={Qin, Yujia and Liang, Shihao and Ye, Yining and Zhu, Kunlun and Yan, Lan and Lu, Yaxi and Lin, Yankai and Cong, Xin and Tang, Xiangru and Qian, Bill and others},
  booktitle={International Conference on Learning Representations},
  volume={2024},
  pages={9695--9717},
  year={2024}
}

@article{patil2024gorilla,
  title={Gorilla: Large language model connected with massive apis},
  author={Patil, Shishir G and Zhang, Tianjun and Wang, Xin and Gonzalez, Joseph E},
  journal={Advances in Neural Information Processing Systems},
  volume={37},
  pages={126544--126565},
  year={2024}
}

@article{zhuang2023toolqa,
  title={Toolqa: A dataset for llm question answering with external tools},
  author={Zhuang, Yuchen and Yu, Yue and Wang, Kuan and Sun, Haotian and Zhang, Chao},
  journal={Advances in Neural Information Processing Systems},
  volume={36},
  pages={50117--50143},
  year={2023}
}

@inproceedings{li2023api,
  title={Api-bank: A comprehensive benchmark for tool-augmented llms},
  author={Li, Minghao and Zhao, Yingxiu and Yu, Bowen and Song, Feifan and Li, Hangyu and Yu, Haiyang and Li, Zhoujun and Huang, Fei and Li, Yongbin},
  booktitle={Proceedings of the 2023 conference on empirical methods in natural language processing},
  pages={3102--3116},
  year={2023}
}

@inproceedings{zhou2024webarena,
  title={Webarena: A realistic web environment for building autonomous agents},
  author={Zhou, Shuyan and Xu, Frank F and Zhu, Hao and Zhou, Xuhui and Lo, Robert and Sridhar, Abishek and Cheng, Xianyi and Ou, Tianyue and Bisk, Yonatan and Fried, Daniel and others},
  booktitle={International Conference on Learning Representations},
  volume={2024},
  pages={15585--15606},
  year={2024}
}

@article{deng2023mind2web,
  title={Mind2web: Towards a generalist agent for the web},
  author={Deng, Xiang and Gu, Yu and Zheng, Boyuan and Chen, Shijie and Stevens, Sam and Wang, Boshi and Sun, Huan and Su, Yu},
  journal={Advances in Neural Information Processing Systems},
  volume={36},
  pages={28091--28114},
  year={2023}
}

@inproceedings{liu2024agentbench,
  title={Agentbench: Evaluating llms as agents},
  author={Liu, Xiao and Yu, Hao and Zhang, Hanchen and Xu, Yifan and Lei, Xuanyu and Lai, Hanyu and Gu, Yu and Ding, Hangliang and Men, Kaiwen and Yang, Kejuan and others},
  booktitle={International Conference on Learning Representations},
  volume={2024},
  pages={52989--53046},
  year={2024}
}

@article{ma2024agentboard,
  title={Agentboard: An analytical evaluation board of multi-turn llm agents},
  author={Ma, Chang and Zhang, Junlei and Zhu, Zhihao and Yang, Cheng and Yang, Yujiu and Jin, Yaohui and Lan, Zhenzhong and Kong, Lingpeng and He, Junxian},
  journal={Advances in neural information processing systems},
  volume={37},
  pages={74325--74362},
  year={2024}
}

@article{ouyang2022training,
  title={Training language models to follow instructions with human feedback},
  author={Ouyang, Long and Wu, Jeffrey and Jiang, Xu and Almeida, Diogo and Wainwright, Carroll and Mishkin, Pamela and Zhang, Chong and Agarwal, Sandhini and Slama, Katarina and Ray, Alex and others},
  journal={Advances in neural information processing systems},
  volume={35},
  pages={27730--27744},
  year={2022}
}

@article{bai2022constitutional,
  title={Constitutional ai: Harmlessness from ai feedback},
  author={Bai, Yuntao and Kadavath, Saurav and Kundu, Sandipan and Askell, Amanda and Kernion, Jackson and Jones, Andy and Chen, Anna and Goldie, Anna and Mirhoseini, Azalia and McKinnon, Cameron and others},
  journal={arXiv preprint arXiv:2212.08073},
  year={2022}
}

@article{yao2022webshop,
  title={Webshop: Towards scalable real-world web interaction with grounded language agents},
  author={Yao, Shunyu and Chen, Howard and Yang, John and Narasimhan, Karthik},
  journal={Advances in Neural Information Processing Systems},
  volume={35},
  pages={20744--20757},
  year={2022}
}

@article{shinn2023reflexion,
  title={Reflexion: Language agents with verbal reinforcement learning},
  author={Shinn, Noah and Cassano, Federico and Gopinath, Ashwin and Narasimhan, Karthik and Yao, Shunyu},
  journal={Advances in neural information processing systems},
  volume={36},
  pages={8634--8652},
  year={2023}
}

@article{madaan2023self,
  title={Self-refine: Iterative refinement with self-feedback},
  author={Madaan, Aman and Tandon, Niket and Gupta, Prakhar and Hallinan, Skyler and Gao, Luyu and Wiegreffe, Sarah and Alon, Uri and Dziri, Nouha and Prabhumoye, Shrimai and Yang, Yiming and others},
  journal={Advances in neural information processing systems},
  volume={36},
  pages={46534--46594},
  year={2023}
}

@inproceedings{zhao2024expel,
  title={Expel: Llm agents are experiential learners},
  author={Zhao, Andrew and Huang, Daniel and Xu, Quentin and Lin, Matthieu and Liu, Yong-Jin and Huang, Gao},
  booktitle={Proceedings of the AAAI Conference on Artificial Intelligence},
  volume={38},
  number={17},
  pages={19632--19642},
  year={2024}
}

@article{zhou2023language,
  title={Language agent tree search unifies reasoning acting and planning in language models},
  author={Zhou, Andy and Yan, Kai and Shlapentokh-Rothman, Michal and Wang, Haohan and Wang, Yu-Xiong},
  journal={arXiv preprint arXiv:2310.04406},
  year={2023}
}

@article{wang2024agent,
  title={Agent workflow memory},
  author={Wang, Zora Zhiruo and Mao, Jiayuan and Fried, Daniel and Neubig, Graham},
  journal={arXiv preprint arXiv:2409.07429},
  year={2024}
}

@article{li2026skillsbench,
  title={SkillsBench: Benchmarking how well agent skills work across diverse tasks},
  author={Li, Xiangyi and Chen, Wenbo and Liu, Yimin and Zheng, Shenghan and Chen, Xiaokun and He, Yifeng and Li, Yubo and You, Bingran and Shen, Haotian and Sun, Jiankai and others},
  journal={arXiv preprint arXiv:2602.12670},
  year={2026}
}

@article{su2026skill,
  title={Skill retrieval augmentation for agentic AI},
  author={Su, Weihang and Long, Jianming and Ai, Qingyao and He, Qiaozhi and Tang, Yichen and Wang, Changyue and Tu, Yiteng and Wang, Yingbo and Liu, Yiqun},
  journal={arXiv preprint arXiv:2604.24594},
  year={2026}
}

@article{zhou2026skillgenbench,
  title={Skillgenbench: Benchmarking skill generation pipelines for llm agents},
  author={Zhou, Yifan and Zhang, Zhentao and Cheng, Ziming and Zhang, Shuo and Lan, Qizhen and Chen, Zhangquan and Yang, Zhi and Chen, Ronghao and Wang, Huacan and Hu, Sen and others},
  journal={arXiv preprint arXiv:2605.18693},
  year={2026}
}

@article{zhong2026skilllearnbench,
  title={SkillLearnBench: Benchmarking Continual Learning Methods for Agent Skill Generation on Real-World Tasks},
  author={Zhong, Shanshan and Lu, Yi and Ning, Jingjie and Wan, Yibing and Feng, Lihan and Ao, Yuyi and Ribeiro, Leonardo FR and Dreyer, Markus and Ammirati, Sean and Xiong, Chenyan},
  journal={arXiv preprint arXiv:2604.20087},
  year={2026}
}

@article{zheng2025skillweaver,
  title={Skillweaver: Web agents can self-improve by discovering and honing skills},
  author={Zheng, Boyuan and Fatemi, Michael Y and Jin, Xiaolong and Wang, Zora Zhiruo and Gandhi, Apurva and Song, Yueqi and Gu, Yu and Srinivasa, Jayanth and Liu, Gaowen and Neubig, Graham and others},
  journal={arXiv preprint arXiv:2504.07079},
  year={2025}
}

@article{ma2026skillgen,
  title={Skillgen: Verified inference-time agent skill synthesis},
  author={Ma, Yuchen and Huang, Yue and Bao, Han and Zhuang, Haomin and Shukla, Swadheen and Galley, Michel and Zhang, Xiangliang and Feuerriegel, Stefan},
  journal={arXiv preprint arXiv:2605.10999},
  year={2026}
}

@article{liu2026skillforge,
  title={Skillforge: Forging domain-specific, self-evolving agent skills in cloud technical support},
  author={Liu, Xingyan and Luo, Xiyue and Li, Linyu and Huang, Ganghong and Liu, Jianfeng and Qiao, Honglin},
  journal={arXiv preprint arXiv:2604.08618},
  year={2026}
}

@article{ma2026skillclaw,
  title={Skillclaw: Let skills evolve collectively with agentic evolver},
  author={Ma, Ziyu and Yang, Shidong and Ji, Yuxiang and Wang, Xucong and Wang, Yong and Hu, Yiming and Huang, Tongwen and Chu, Xiangxiang},
  journal={arXiv preprint arXiv:2604.08377},
  year={2026}
}

@article{yang2026skillopt,
  title={Skillopt: Executive strategy for self-evolving agent skills},
  author={Yang, Yifan and Gong, Ziyang and Huang, Weiquan and Yang, Qihao and Zhou, Ziwei and Huang, Zisu and Li, Yan and Gao, Xuemei and Dai, Qi and Liu, Bei and others},
  journal={arXiv preprint arXiv:2605.23904},
  year={2026}
}

@article{wu2023autogen,
  title={Autogen: Enabling next-gen llm applications via multi-agent conversation},
  author={Wu, Qingyun and Bansal, Gagan and Zhang, Jieyu and Wu, Yiran and Li, Beibin and Zhu, Erkang and Jiang, Li and Zhang, Xiaoyun and Zhang, Shaokun and Liu, Jiale and others},
  journal={arXiv preprint arXiv:2308.08155},
  year={2023}
}

@article{yang2024swe,
  title={Swe-agent: Agent-computer interfaces enable automated software engineering},
  author={Yang, John and Jimenez, Carlos and Wettig, Alexander and Lieret, Kilian and Yao, Shunyu and Narasimhan, Karthik and Press, Ofir},
  journal={Advances in Neural Information Processing Systems},
  volume={37},
  pages={50528--50652},
  year={2024}
}

@article{yao2024tau,
  title={Tau-Bench: A Benchmark for Tool-Agent-User Interaction in Real-World Domains},
  author={Yao, Shunyu and Shinn, Noah and Razavi, Pedram and Narasimhan, Karthik},
  journal={arXiv preprint arXiv:2406.12045},
  year={2024}
}

@article{xie2024travelplanner,
  title={Travelplanner: A benchmark for real-world planning with language agents},
  author={Xie, Jian and Zhang, Kai and Chen, Jiangjie and Zhu, Tinghui and Lou, Renze and Tian, Yuandong and Xiao, Yanghua and Su, Yu},
  journal={arXiv preprint arXiv:2402.01622},
  year={2024}
}

@article{wang2023voyager,
  title={Voyager: An open-ended embodied agent with large language models},
  author={Wang, Guanzhi and Xie, Yuqi and Jiang, Yunfan and Mandlekar, Ajay and Xiao, Chaowei and Zhu, Yuke and Fan, Linxi and Anandkumar, Anima},
  journal={arXiv preprint arXiv:2305.16291},
  year={2023}
}

@misc{qwen3technicalreport,
      title={Qwen3 Technical Report}, 
      author={{Qwen Team}},
      year={2025},
      eprint={2505.09388},
      archivePrefix={arXiv},
      primaryClass={cs.CL},
      url={https://arxiv.org/abs/2505.09388}, 
}

@misc{qwen3.5,
    title  = {{Qwen3.5}: Towards Native Multimodal Agents},
    author = {{Qwen Team}},
    month  = {February},
    year   = {2026},
    url    = {https://qwen.ai/blog?id=qwen3.5}
}

@article{xu2026deepseek,
  title={Deepseek-v4: Towards highly efficient million-token context intelligence},
  author={Xu, Anyi and Lin, Bangcai and Xue, Bing and Wang, Bingxuan and Xu, Bingzheng and Wu, Bochao and Zhang, Bowei and Lin, Chaofan and Dong, Chen and Ling, Chenchen and others},
  journal={arXiv preprint arXiv:2606.19348},
  year={2026}
}

@misc{google2026gemini31flashlite,
  title        = {{Gemini 3.1 Flash-Lite}: Built for Intelligence at Scale},
  author       = {{Gemini Team}},
  year         = {2026},
  month        = {March},
  url          = {https://blog.google/innovation-and-ai/models-and-research/gemini-models/gemini-3-1-flash-lite/}
}

@misc{openai2024gpt4omini,
  title        = {{GPT-4o mini}: Advancing Cost-Efficient Intelligence},
  author       = {{OpenAI}},
  year         = {2024},
  month        = {July},
  url          = {https://openai.com/index/gpt-4o-mini-advancing-cost-efficient-intelligence/}
}

@misc{anthropic2026claudeopus46,
  title        = {{Claude Opus 4.6}},
  author       = {{Anthropic}},
  year         = {2026},
  month        = {February},
  url          = {https://www.anthropic.com/news/claude-opus-4-6}
}

@software{Ding_WildClawBench,
  author  = {Ding, Shuangrui and Dai, Xuanlang and Xing, Long and Ding, Shengyuan and Liu, Ziyu and Yang, Jingyi and Yang, Penghui and Zhang, Zhixiong and Wei, Xilin and Ma, Yubo and Duan, Haodong and Shao, Jing and Wang, Jiaqi and Lin, Dahua and Chen, Kai and Zang, Yuhang},
  title   = {WildClawBench},
  year    = {2026},
  url     = {https://github.com/InternLM/WildClawBench},
  license = {MIT}
}

@article{yang2026autoskill,
  title={Autoskill: Experience-driven lifelong learning via skill self-evolution},
  author={Yang, Yutao and Li, Junsong and Pan, Qianjun and Zhan, Bihao and Cai, Yuxuan and Du, Lin and Zhou, Jie and Chen, Kai and Chen, Qin and Li, Xin and others},
  journal={arXiv preprint arXiv:2603.01145},
  year={2026}
}

@inproceedings{parmar2024logicbench,
  title={Logicbench: Towards systematic evaluation of logical reasoning ability of large language models},
  author={Parmar, Mihir and Patel, Nisarg and Varshney, Neeraj and Nakamura, Mutsumi and Luo, Man and Mashetty, Santosh and Mitra, Arindam and Baral, Chitta},
  booktitle={Proceedings of the 62nd Annual Meeting of the Association for Computational Linguistics (Volume 1: Long Papers)},
  pages={13679--13707},
  year={2024}
}

@inproceedings{reimers-2019-sentence-bert,
    title = "Sentence-BERT: Sentence Embeddings using Siamese BERT-Networks",
    author = "Reimers, Nils and Gurevych, Iryna",
    booktitle = "Proceedings of the 2019 Conference on Empirical Methods in Natural Language Processing",
    month = "11",
    year = "2019",
    publisher = "Association for Computational Linguistics",
    url = "https://arxiv.org/abs/1908.10084",
}

@article{zhang2026self,
  title={Self-harness: Harnesses that improve themselves},
  author={Zhang, Hangfan and Zhang, Shao and Li, Kangcong and Zhang, Chen and Chen, Yang and Zhang, Yiqun and Bai, Lei and Hu, Shuyue},
  journal={arXiv preprint arXiv:2606.09498},
  year={2026}
}

@article{gao2026skillaudit,
  title={SkillAudit: Ground-Truth-Free Skill Evolution via Paired Trajectory Auditing},
  author={Gao, Haowen and Chen, Haoran and Wang, Can and Guo, Shasha and Pang, Liang and Liu, Zhaoyang and Shen, Huawei and Cheng, Xueqi},
  journal={arXiv preprint arXiv:2606.14239},
  year={2026}
}


\end{document}